\documentclass[lettersize,journal]{IEEEtran}
\usepackage{amsmath,amssymb,amsfonts}
\usepackage{algorithmic}
\usepackage{algorithm}
\usepackage{array}
\usepackage{graphicx}
\usepackage[caption=false,font=footnotesize,labelfont=rm,textfont=rm]{subfig}
\usepackage{textcomp}
\usepackage{stfloats}
\usepackage{booktabs}
\usepackage{subfig}
\usepackage{subfloat}
\usepackage{multirow}
\usepackage{multicol}
\usepackage{stfloats}
\usepackage{siunitx}
\usepackage{url}
\usepackage{verbatim}
\usepackage{graphicx}
\usepackage{amsmath}
\usepackage{bbding}
\usepackage[algo2e]{algorithm2e}
\usepackage{cite}
\usepackage{tcolorbox}
\usepackage{bbding}

\usepackage{tcolorbox}
\tcbuselibrary{skins,breakable}
\usepackage{listings}
\usepackage{xcolor}
\usepackage{enumitem}

\usepackage{makecell}
\usepackage{booktabs}
\usepackage{colortbl}

\usepackage{pifont}

\lstdefinestyle{promptjson}{
    basicstyle=\ttfamily\scriptsize,
    breaklines=true,
    breakatwhitespace=false,
    columns=fullflexible,
    keepspaces=true,
    showstringspaces=false,
    frame=none,
    backgroundcolor=\color{gray!5},
    xleftmargin=0.5em,
    xrightmargin=0.5em
}

\newcommand{\cmark}{\CheckmarkBold}
\newcommand{\xmark}{\XSolidBrush}

\usepackage{xcolor}
\usepackage{enumitem}
\usepackage{tcolorbox}
\tcbuselibrary{breakable}

\newtcolorbox{promptbox}[1]{
    breakable,
    colback=white,
    colframe=gray!75!black,
    colbacktitle=gray!75!black,
    coltitle=white,
    title={#1},
    fonttitle=\large\bfseries,
    boxrule=0.8pt,
    arc=2mm,
    left=8pt,
    right=8pt,
    top=6pt,
    bottom=6pt,
    toptitle=6pt,
    bottomtitle=6pt,
    width=\textwidth
}

\usepackage{threeparttable} 
\begin{document}

\title{AffectSeek: Agentic Affective Understanding in Long Videos under Vague User Queries}

\author{Zhen Zhang, Yuhang Yang, Yunxiang Jiang, Yuhuan Lu, Haifeng Lu, \\Zheng Lian, Runhao Zeng$^{*}$, and Xiping Hu$^{*}$
\thanks{ $*$ Corresponding author}
\thanks{Zhen Zhang is with Gansu Provincial Key Laboratory of
Wearable Computing, School of Information Science and Engineering,
Lanzhou University, Gansu 730000, China, and Guangdong-Hong Kong-Macao Joint Laboratory for Emotional Intelligence and Pervasive Computing, Shenzhen MSU-BIT University, Shenzhen 518107, China (e-mail: zhangzhen19@lzu.edu.cn).}
\thanks{Yuhuan Lu is with the Department of Computer and Information Engineering, Khalifa University, Abu Dhabi 127788, United Arab Emirates. (e-mail: yc17462@connect.um.edu.mo)}
\thanks{Haifeng Lu is with the Artificial Intelligence Research Institute, Shenzhen MSU-BIT University, Shenzhen 518107, China, and the Department of Electrical and Computer Engineering, The University of Hong Kong, China (e-mail: luhfhku@hku.hk)}
\thanks{Zheng Lian is with Shanghai Reasearch Institute for Intelligent Autonomous Systems, Tongji University, Shanghai 201210, China (e-mail: lianzheng@tongji.edu.cn)}
\thanks{Xiping Hu, Runhao Zeng, Yuhang Yang and Yuxiang Jiang are with the Artificial Intelligence Research Institute, Shenzhen MSU-BIT University and Guangdong-Hong Kong-Macao Joint Laboratory for Emotional Intelligence and Pervasive Computing, Shenzhen MSU-BIT University, Shenzhen 518107, China. (e-mail: huxp, zengrh@smbu.edu.cn).}}

\markboth{Journal of \LaTeX\ Class Files,~Vol.~14, No.~8, August~2021}%
{Shell \MakeLowercase{\textit{et al.}}: A Sample Article Using IEEEtran.cls for IEEE Journals}


\maketitle

\begin{abstract}
Existing affective understanding studies have mainly focused on recognizing emotions from images, audio signals, or pre-cliped video clips, where the affective evidence is already given. This passive and clip-centered setting does not fully reflect real-world scenarios, in which users often interact with long videos and express their needs through natural-language queries. In this paper, we study \textbf{Vague-Query-driven video Affective Understanding (VQAU)}, a new task that requires models to localize affective moments in long videos, predict their emotion categories, and generate evidence-grounded rationales under vague user queries. To support this task, we construct \textbf{VQAU-Bench}, a benchmark that integrates long videos, vague affective queries, temporal clip annotations, emotion labels, and rationale explanations into a unified evaluation framework. VQAU-Bench enables systematic assessment of semantic-temporal-affective alignment, affective moment localization, emotion classification, and rationale generation. To address the multi-step reasoning challenges of VQAU, we further propose \textbf{AffectSeek}, an agentic framework that actively seeks, verifies, and explains affective moments in long videos. AffectSeek decomposes VQAU into intent interpretation, candidate localization, clip verification, emotion reasoning, and rationale generation, and progressively aligns vague user intent with long-video evidence through role-specialized reasoning and cross-stage verification. Experiments show that VQAU remains challenging for existing affective recognition models and single-step vision-language models, while AffectSeek provides a simple yet effective framework for agentic long-video affective understanding.
\end{abstract}


\section{Introduction}
\label{sec:introduction}

\begin{figure}[h!]
  \centering
  \includegraphics[width=1.0\linewidth]{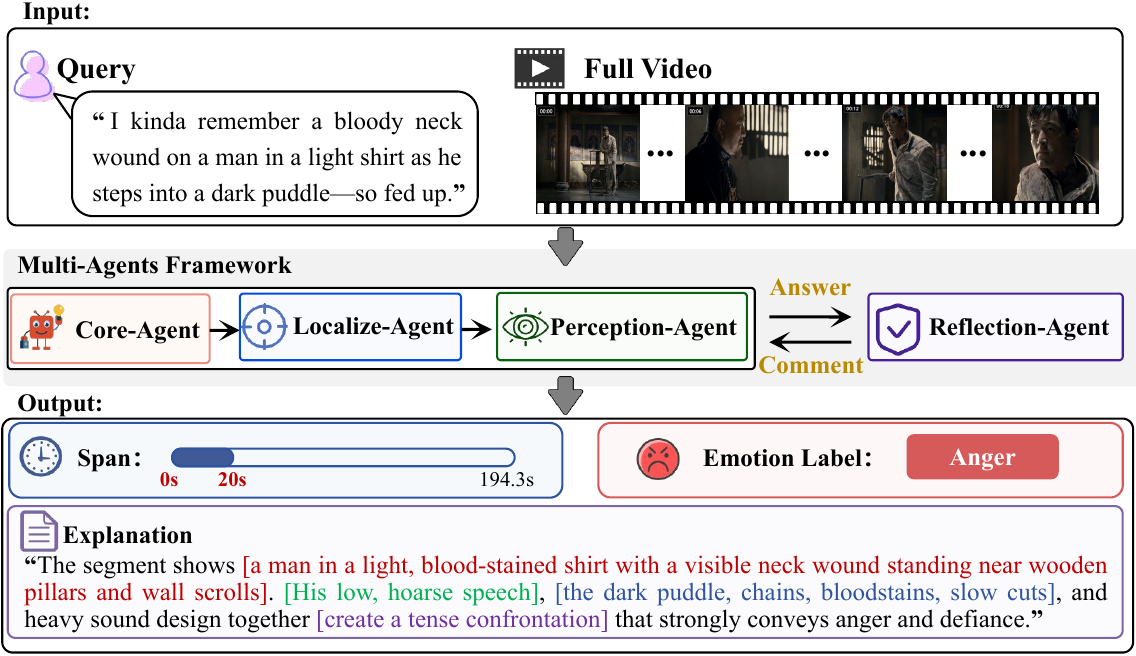}
  \caption{VQAU task-solving pipeline. Given a query and a full video as inputs, our multi-agent framework predicts the temporal location of the affective event, identifies its emotion category, and generates the corresponding explanatory rationale.}
  \label{fig:task}
\end{figure}

\IEEEPARstart{A}{ffective} understanding, which aims to infer human emotional states from multimodal signals, has become a central research topic in affective computing, with broad applications in human-computer interaction, mental health analysis, educational assistance, and social media understanding~\cite{wu2025multimodal, shangguan2025facial, lu2025emotion}. Recent studies have achieved substantial progress in facial expression recognition, speech emotion recognition, multimodal affective fusion, and video emotion classification~\cite{10766406, mao2025poster++, khan2024mser, CHEN2026113367, zhu2025rmer, Lu_Chen_Liang_Tan_Zeng_Hu_2025, 11442664, zhang2025skeleton}. Despite these advances, most existing affective understanding tasks remain passive and clip-centered. They usually assume that the affective evidence has already been selected and provided to the model in the form of an image, an audio clip, or a pre-cliped video clip. Under this setting, the model is mainly required to map the given input to an emotion category, rather than to discover where affective evidence occurs in a long video, whether it matches a user's intent, and why it supports a particular emotional interpretation.

However, this conventional setting does not fully reflect the complexity of real-world affective understanding. In practical video analysis and content creation scenarios, users often interact with long videos rather than precisely edited affective clips~\cite{wang2025long, luo2023joint}. A long video may contain multiple affective moments, complex narrative contexts, and continuously evolving emotional states. Moreover, users usually express their needs through natural-language queries rather than predefined emotion labels or manually specified temporal boundaries. Therefore, a more realistic affective understanding system should be able to interpret user intent, localize target affective moments in long videos, infer their emotion categories, and provide evidence-grounded explanations.

Existing benchmarks are insufficient for systematically evaluating these abilities. Traditional affective datasets mainly provide emotion labels for images~\cite{li2017reliable,mollahosseini2017affectnet}, audio samples~\cite{burkhardt2000database,lotfian2017building}, or pre-cropped video clips~\cite{lian2025affectgpt}, and thus do not require models to localize affective content from long videos. Some recent datasets introduce multimodal affective descriptions or temporal annotations, but they often rely on explicit events, fixed-form queries, or already provided affective clips~\cite{zhang2022temporal,vicol2018moviegraphs}. More importantly, existing datasets rarely provide long-video inputs, user queries, temporal clip boundaries, emotion category labels, and evidence-grounded rationales in a unified form. As a result, they cannot effectively evaluate whether a model can establish reliable semantic-temporal-affective correspondences between user intent and long-video evidence. This motivates us to construct a new benchmark for user-centered long-video affective understanding.

To this end, we introduce a new task: \textbf{\underline{V}}ague-\textbf{\underline{Q}}uery-driven video \textbf{\underline{A}}ffective \textbf{\underline{U}}nderstanding (VQAU), as shown in Fig.~\ref{fig:task}. Given a long video and a natural-language user query, the model is required to localize the affective moment corresponding to the query, recognize its emotion category, and generate an evidence-grounded rationale that supports both the localization and the emotion prediction. To make the task closer to real user interactions and more challenging for model evaluation, we instantiate the query as a vague affective query rather than an explicit event description, a predefined emotion label, or a manually specified temporal boundary. Such a query reflects a user's partial and subjective impression of an affective moment, such as a relieved expression, a comforting atmosphere, or a moment where music, scene context, and character behavior jointly convey an emotion. Therefore, VQAU requires the model not only to match textual semantics with video content, but also to infer latent affective intent and ground it in temporally distributed multimodal evidence.

Based on this task formulation, we construct \textbf{VQAU-Bench}, a new benchmark for vague-query-driven affective understanding in long videos. Each sample in VQAU-Bench consists of a long video, a vague affective user query, the corresponding temporal clip annotation, an emotion category label, and an evidence-grounded rationale explanation. This annotation design supports the evaluation of three core abilities: affective moment localization, emotion classification, and rationale generation. Compared with existing affective datasets, VQAU-Bench integrates long-video input, vague affective user queries, temporal grounding, emotion labels, and explanatory rationales into a unified benchmark, providing a data foundation for studying user-centered affective understanding beyond passive clip-level recognition.

VQAU is substantially more complex than conventional clip-level emotion recognition and general video understanding. The model needs to interpret vague affective intent, search for sparse evidence across a long video, verify whether candidate moments are consistent with the query, reason about emotions from multimodal cues, and generate explanations grounded in the selected clip. These steps are tightly coupled: an incorrect localization may lead to an incorrect emotion prediction, while an unsupported emotion judgment may produce a plausible but ungrounded explanation. Therefore, directly applying traditional affective recognition models or single-step vision-language models is often insufficient for this setting. These challenges suggest the need for an agentic affective understanding framework, where the model acts not merely as a passive emotion classifier, but as an active reasoning system that can interpret user intent, seek and verify evidence, reason about emotions, and explain its decisions.

To address this challenge, we propose \textbf{AffectSeek}, a simple yet effective agentic framework for VQAU. As its name suggests, AffectSeek actively seeks, verifies, and explains affective moments in long videos under vague user queries. Instead of directly predicting the final answer in one step, AffectSeek decomposes VQAU into several role-specialized stages, including query understanding, candidate moment localization, clip verification, emotion reasoning, evidence assessment, and rationale generation. The key idea is not merely to divide the task into sequential modules, but to introduce cross-stage verification among localization, emotion prediction, and evidence grounding. Candidate moments are first retrieved according to vague affective cues, then checked against multimodal video evidence and emotional consistency before the final category and rationale are produced. In this way, AffectSeek progressively aligns vague user queries with long-video evidence and improves the reliability and interpretability of affective understanding.

Overall, this paper studies affective understanding under a more realistic long-video and user-query-driven setting, where models are required to localize, recognize, and explain affective moments rather than classify pre-segmented clips. The main contributions of this paper are summarized as follows:

\begin{itemize}
    \item We introduce \textbf{VQAU}, a new task for vague-query-driven affective understanding in long videos, which requires models to jointly perform affective moment localization, emotion recognition, and evidence-grounded rationale generation.

    \item We construct \textbf{VQAU-Bench}, a benchmark that unifies long videos, vague affective user queries, temporal clip annotations, emotion labels, and rationale explanations, enabling systematic evaluation of semantic-temporal-affective alignment.

    \item We propose \textbf{AffectSeek}, an agentic framework that actively seeks, verifies, and explains affective moments in long videos under vague user queries, improving the reliability and interpretability of VQAU.
\end{itemize}

\section{Related Works}
\label{relatedworks}

\subsection{Clip-Level Affective Understanding}

Recent studies on multimodal affective understanding have gradually shifted from conventional emotion classification to fine-grained emotion recognition and explanation with multimodal large models. Emotion-LLaMA \cite{cheng2024emotion} explores multimodal affective understanding by integrating audio, visual, and textual cues through instruction tuning. AffectGPT \cite{lian2025affectgpt} further introduces open-vocabulary emotion recognition and emphasizes natural-language-based affective understanding in open environments. Emotion-LLaMA-V2 \cite{peng2026emotion} advances this research direction by introducing an end-to-end multi-view encoder, a convolutional attention pre-fusion module, and a perception-to-cognition curriculum instruction learning strategy. In addition, Agent-MER \cite{DBLP:conf/mm/Lai0H025} formulates multimodal emotion recognition as a knowledge-guided cognitive-agent reasoning process and improves fine-grained emotion recognition through hierarchical deliberation and self-consistency voting.

Despite these advances, most existing affective understanding methods are still built on predefined or manually selected emotional clips. Their primary goal is to recognize or explain the emotion contained in a given image, utterance, or video clip, rather than to identify where affective content occurs in an original video. This setting is insufficient for realistic user-centered scenarios, where users usually do not provide precise temporal boundaries. Instead, they often describe vague memories, subjective impressions, or emotional experiences, and expect the system to retrieve the corresponding affective event from a long or untrimmed video. Therefore, existing clip-level affective understanding methods cannot fully address the problem studied in this paper, which requires models to jointly localize affective clips, classify emotions, and provide evidence-grounded explanations from original videos.

\begin{figure}[t]
  \centering
  \includegraphics[width=1.0\linewidth]{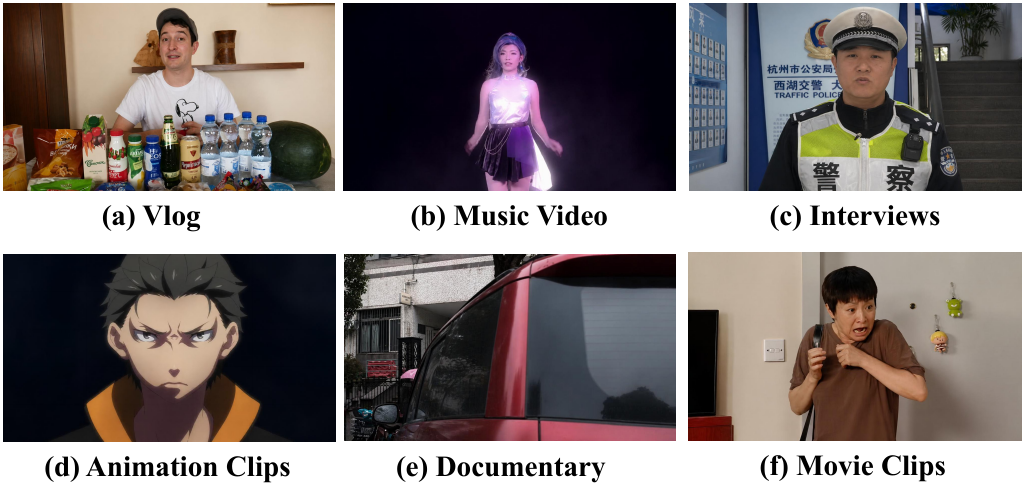}
  \caption{Illustration of partial dataset scenarios. Specifically, (a)–(f) show the scenario examples included in the dataset.}
  \label{fig:scene}
\end{figure}

\subsection{Query-Guided Affective Event Discovery}

A few recent studies have begun to incorporate temporal localization and reasoning into affective video understanding. For example, Hawkeye \cite{zhao2024hawkeye} focuses on discovering and localizing implicit anomalous sentiment in reconnaissance videos under fixed language instructions. Omni-SILA \cite{luo2025omni} leverages implicit scene information, including actions, object relations, and background context, to identify, localize, and attribute positive and negative emotions in videos. EmoDETective \cite{huang2025emodetective} further studies emotional cause detection and reasoning within video clips. These works indicate that affective understanding is moving beyond isolated classification toward more comprehensive localization and reasoning.

However, existing affective localization studies still differ substantially from the task considered in this work. Hawkeye and Omni-SILA mainly rely on predefined task instructions and are designed for specific sentiment-oriented settings, while EmoDETective focuses on emotion analysis and causal reasoning over given video content. None of these works addresses the more practical scenario in which a user provides a vague natural-language query as guidance, and the model must first discover the target affective event from an original video before performing emotion classification and rationale analysis. In contrast, our work studies affective video understanding under user-centered vague queries, requiring models to establish a reliable correspondence among subjective language descriptions, temporal video evidence, emotion labels, and explanatory rationales. This setting better reflects real-world multimodal interaction and exposes limitations that cannot be sufficiently evaluated by existing clip-level affective understanding benchmarks.

\begin{figure}[t]
  \centering
  \includegraphics[width=0.9\linewidth]{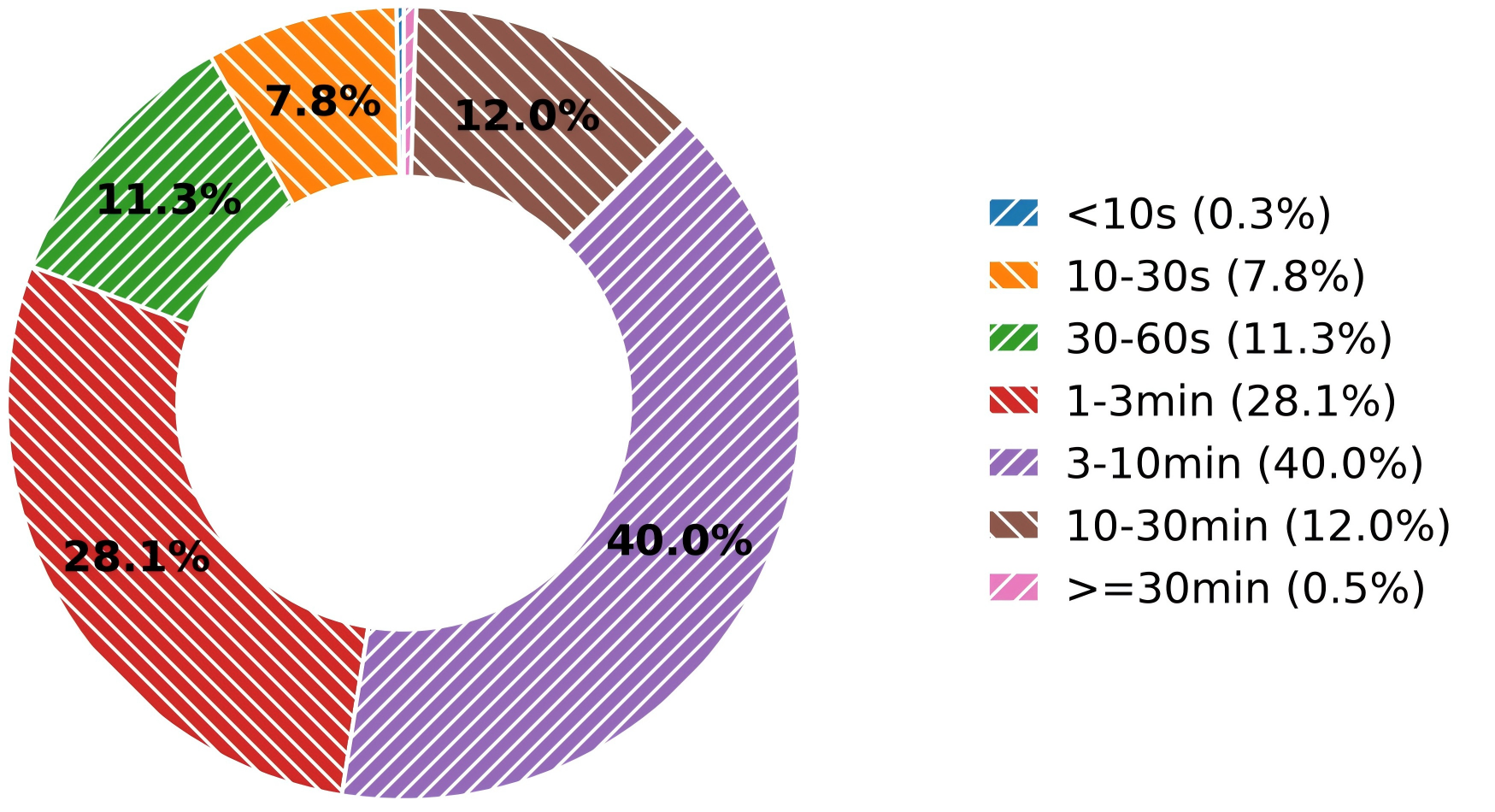}
  \caption{Distribution of video durations in the dataset. Most full-length videos are between 3 and 10 minutes long.}
  \label{fig:distribution}
\end{figure}

\section{VQAU-Bench: Dataset Construction}
\label{VQAU-Bench}
To better reflect the needs of affective understanding in real-world scenarios and to facilitate research on vague-language-guided affective video understanding, we construct a new multimodal benchmark dataset, VQAU-Bench. Existing emotion datasets mainly focus on emotion classification over pre-segmented clips. Although such datasets are useful for evaluating a model’s emotion recognition capability, they are insufficient for assessing whether a model can, in realistic settings, localize target affective events in raw videos from vague natural-language descriptions and further perform emotion classification and rationale analysis. VQAU-Bench is designed to fill this gap.

Specifically, VQAU-Bench is built upon the multimodal clip-level video data from VAD \cite{wang2024vad}, which was originally developed for emotion category classification. By revisiting the original full videos, we re-analyze and refine the temporal boundaries and emotion categories of event-level affective clips, and further enrich them with rationale explanation annotations. In total, the dataset contains \textbf{2,976} raw videos collected from Bilibili and \textbf{16,292} affective event clips. For each event clip, we design three personalized retrieval-oriented queries, resulting in \textbf{48,876} query--clip pairs in total and covering \textbf{13} emotion categories: Angry, Anticipation, Disgust, Fear, Horror, Joy, Love, Neutral, Pride, Sadness, Satisfaction, Surprise, and Trust.
For rationale explanation annotation, we adopt a human-in-the-loop annotation pipeline. Specifically, we first use a large language model to generate initial explanations, then apply rule-based validation to filter the generated results, and finally ask four graduate students specializing in affective computing to manually review and correct invalid samples. This process improves annotation quality and consistency while maintaining efficiency.

\subsection{Data Collection and Affective Event Temporal Annotation}
The raw video data of VQAU-Bench are sourced from the Bilibili platform and comprise 2,976 lifestyle-oriented videos. Overall, these videos cover a diverse set of real-world scenarios, including vlogs, documentaries, interviews, animation clips, movie clips, and music videos, as shown in Fig. \ref{fig:scene}(a)--(f). The videos have an average duration of 293.21 s, and the detailed distribution of video durations is shown in Fig. \ref{fig:distribution}. Their frame rates primarily range from 25 to 30 FPS. All videos preserve both the original visual content and the corresponding audio, thereby providing a complete visual and acoustic basis for subsequent multimodal affective understanding.

On this basis, we inherit and organize the event-level affective clips from the full videos by leveraging the original annotations in the VAD dataset, ultimately obtaining 16,292 event clips. We retain their original emotion labels, which cover 13 emotion categories: \textbf{Anger, Anticipation, Disgust, Fear, Horror, Joy, Love, Neutral, Pride, Sadness, Satisfaction, Surprise, and Trust} \cite{wang2024vad}. In addition, based on the affective event clips provided by VAD, we identify their corresponding start and end timestamps in the original videos and use them as temporal localization annotations for affective events. These event-level annotations provide the foundation for subsequent data reorganization, query construction, and affective clip re-annotation.

\begin{figure*}[ht!]
  \centering
  \includegraphics[width=1.0\linewidth]{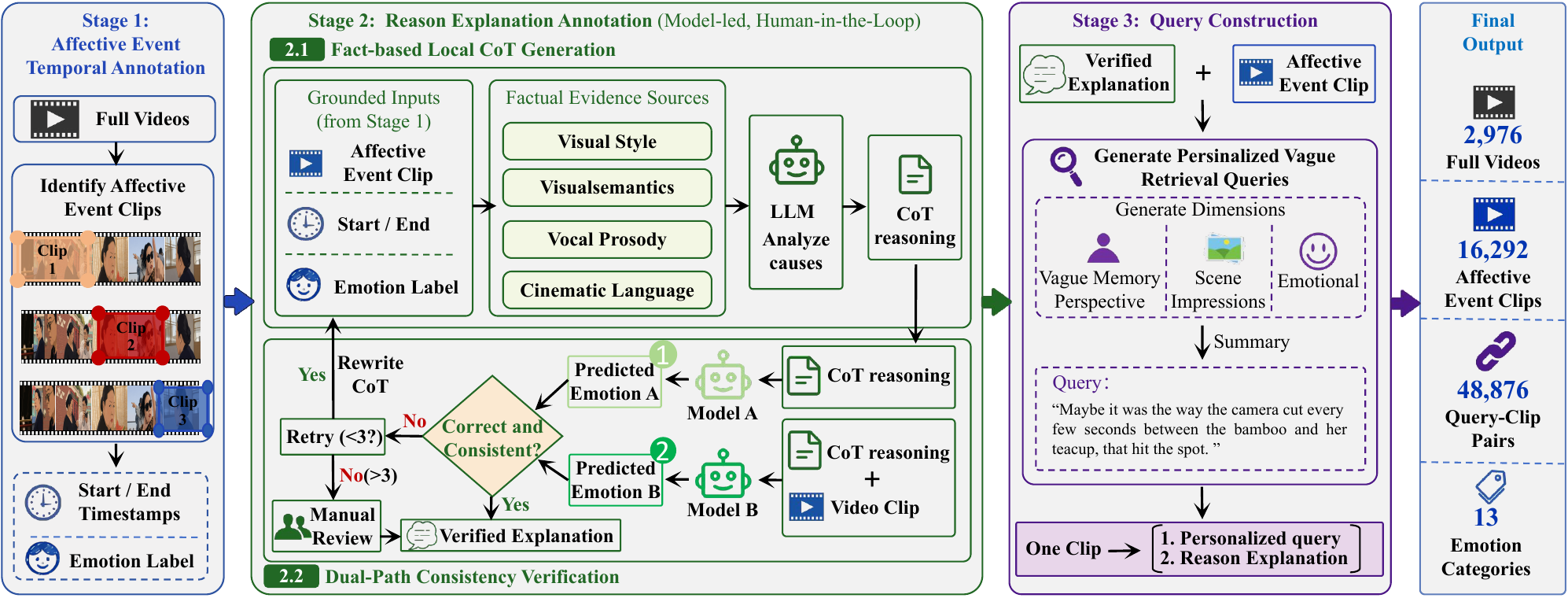}
  \caption{Overview of the proposed annotation pipeline. The framework identifies affective event clips with temporal boundaries and emotion labels, generates and verifies fact-based local CoT explanations, and constructs personalized vague retrieval queries. The final dataset contains 2,976 full videos, 16,292 affective event clips, 48,876 query--clip pairs, and 13 emotion categories.}
  \label{fig:annotation}
\end{figure*}

\subsection{Reason Explanation and Query Construction}

To construct event-level affective annotations with temporal grounding and natural-language retrieval queries, we design a three-stage annotation framework, as shown in Fig.~\ref{fig:annotation}. The framework first localizes affective event clips from full videos, then generates and verifies rationale explanations for each clip, and finally constructs personalized vague retrieval queries based on the verified explanations and video evidence. The detailed prompts are provided in the Appendix A.

\textbf{Stage 1: Affective Event Temporal Annotation.}
Given a collection of full videos, we first identify affective event clips that contain clear emotional expressions or affective scene changes. Each selected clip is annotated with its start and end timestamps, forming a temporally localized affective event clip. Meanwhile, an emotion label is assigned to each clip according to its dominant affective category, following the protocol of Wang et al. \cite{wang2024vad}. This stage provides the grounded inputs for subsequent explanation generation, including the affective event clip, temporal boundaries, and emotion label.

\textbf{Stage 2: Reason Explanation Annotation.}
Based on the temporally localized affective events obtained from Stage 1, we perform model-led, human-in-the-loop reason explanation annotation. This stage consists of two sub-stages: fact-based local CoT generation and dual-path consistency verification.

\textit{(1) Fact-based Local CoT Generation.}
For each affective event clip, we provide the LLM with grounded inputs, including the event clip, its start and end timestamps, and the annotated emotion label. To ensure that the generated explanation is supported by observable video evidence, the model is prompted to analyze the emotional cause from four factual evidence sources: visual style, visual semantics, vocal prosody, and cinematic language. The model then produces a local chain-of-thought (CoT) explanation describing why the clip expresses the annotated emotion. Unlike global video-level summaries, this local CoT is explicitly constrained by the localized event clip and is intended to capture segment-specific affective evidence.

\textit{(2) Dual-Path Consistency Verification.}
To improve the reliability of the generated CoT explanations, we further introduce a dual-path verification mechanism. Specifically, each generated CoT is evaluated through two complementary paths. In the first path, only the CoT explanation is used as input, and a model is asked to infer the corresponding emotion category. In the second path, both the CoT explanation and the original affective event clip are provided as inputs for emotion recognition. A CoT explanation is accepted only when the predictions from both paths are consistent with each other and match the ground-truth emotion label. If the predictions are incorrect or inconsistent, the CoT is rewritten and verified again. This retry process is repeated for at most three rounds. Samples that still fail the consistency check after three iterations are transferred to manual review, where trained annotators inspect the video clip, emotion label, and explanation content to correct potential errors. The accepted or manually corrected explanations are then used as verified rationale explanations.

\textbf{Stage 3: Query Construction.}
After obtaining the validated rationales, we construct personalized vague retrieval queries for each affective event clip. Specifically, the validated rationale and the corresponding event clip are jointly used as the basis for query generation. This process aims to simulate how users describe remembered video clips in real-world retrieval scenarios, where queries are often subjective, incomplete, and impression-based rather than precise factual descriptions. Accordingly, the generated queries are designed to reflect three aspects: vague memory, scene impressions, and emotional experiences.

For each affective event clip, we generate three queries with different forms of expression, such as descriptions beginning with “I think...”, “Maybe it was...”, or “I kind of...”. Each query is paired with the corresponding event clip and its validated rationale, thereby forming query--clip pairs that support affective video localization and explanation evaluation.
Finally, we randomly sample the generated queries and ask four annotators to cross-check them. The annotators evaluate whether each query is relevant to the clip content and affective expression, and whether it is consistent with the actual video content.

Through this annotation pipeline, we obtain \textbf{2,976} full videos, \textbf{16,292} affective event clips, \textbf{48,876} query--clip pairs, and \textbf{13} emotion categories. The final annotations include temporally grounded affective event clips, personalized vague retrieval queries, and verified rationale explanations.


\begin{figure}[t]
  \centering
  \includegraphics[width=1.0\linewidth]{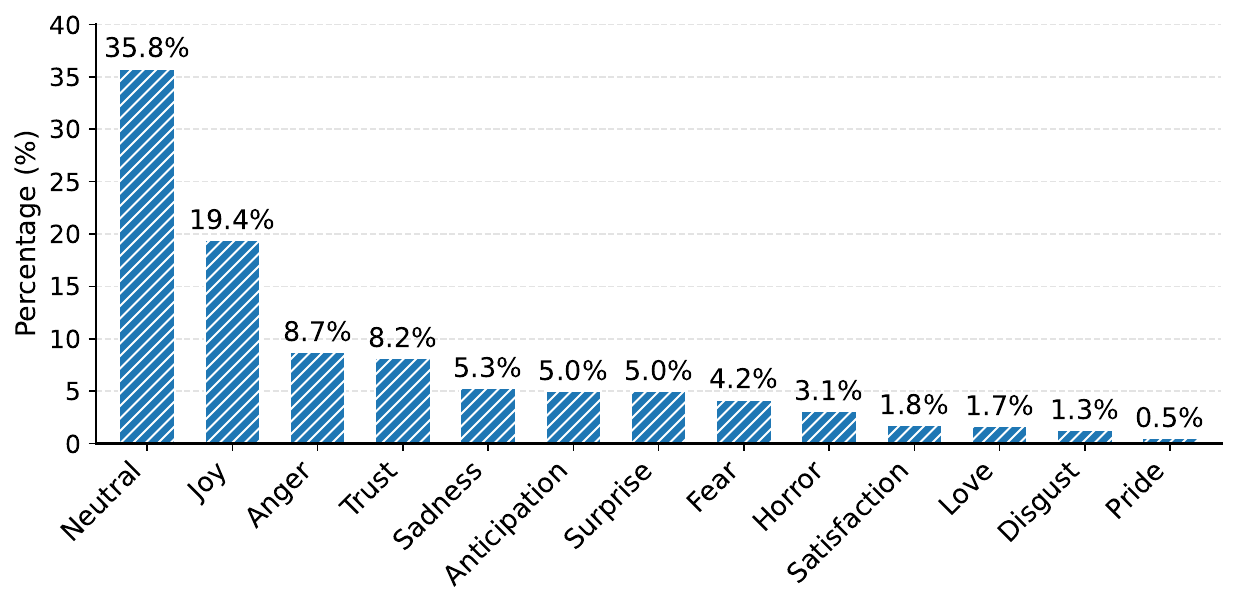}
  \caption{Distribution of samples across emotion categories in the dataset. The dataset covers 13 emotion categories and exhibits a long-tailed distribution, with Neutral and Joy representing the largest proportions.}
  \label{fig:statistics}
\end{figure}

\begin{figure*}[ht!]
  \centering
  \begin{minipage}[b]{0.2\linewidth}
    \centering
    {\small (a) Query Length Donut}\par
    \vspace{0.3em}
    \includegraphics[width=\linewidth]{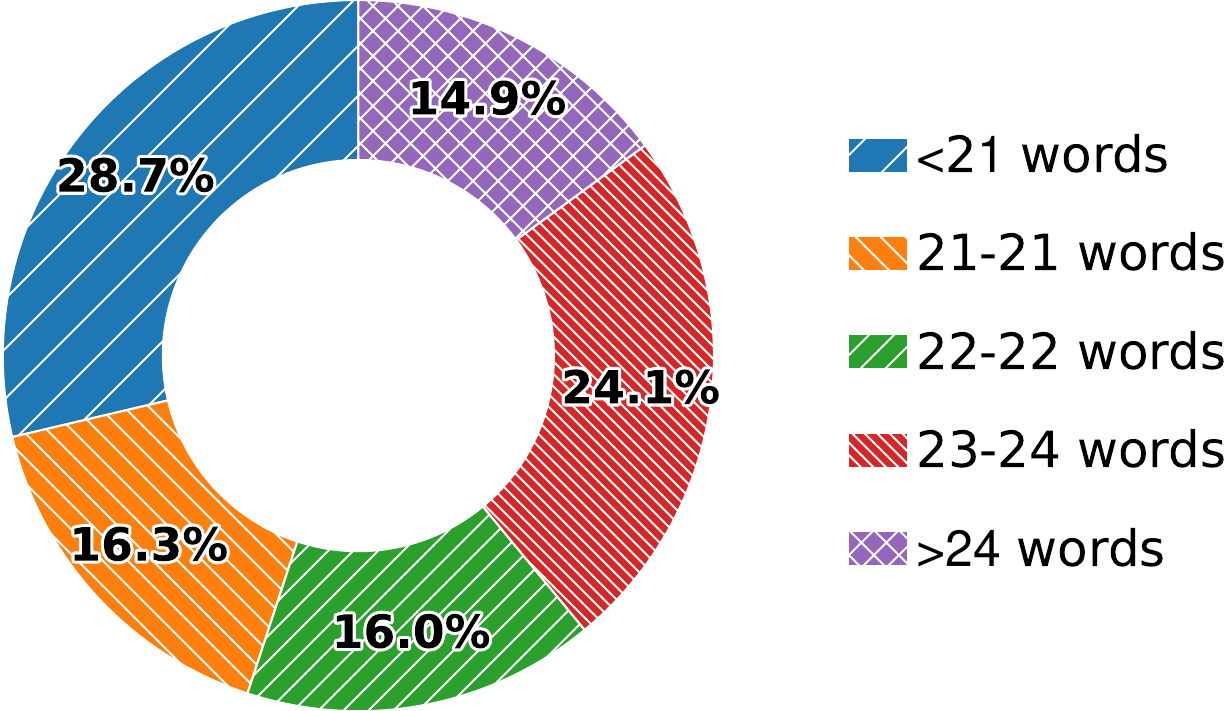}
  \end{minipage}
  \hfill 
  \begin{minipage}[b]{0.215\linewidth}
    \centering
    {\small (b) Reason Length Donut}\par
    \vspace{0.3em}
    \includegraphics[width=\linewidth]{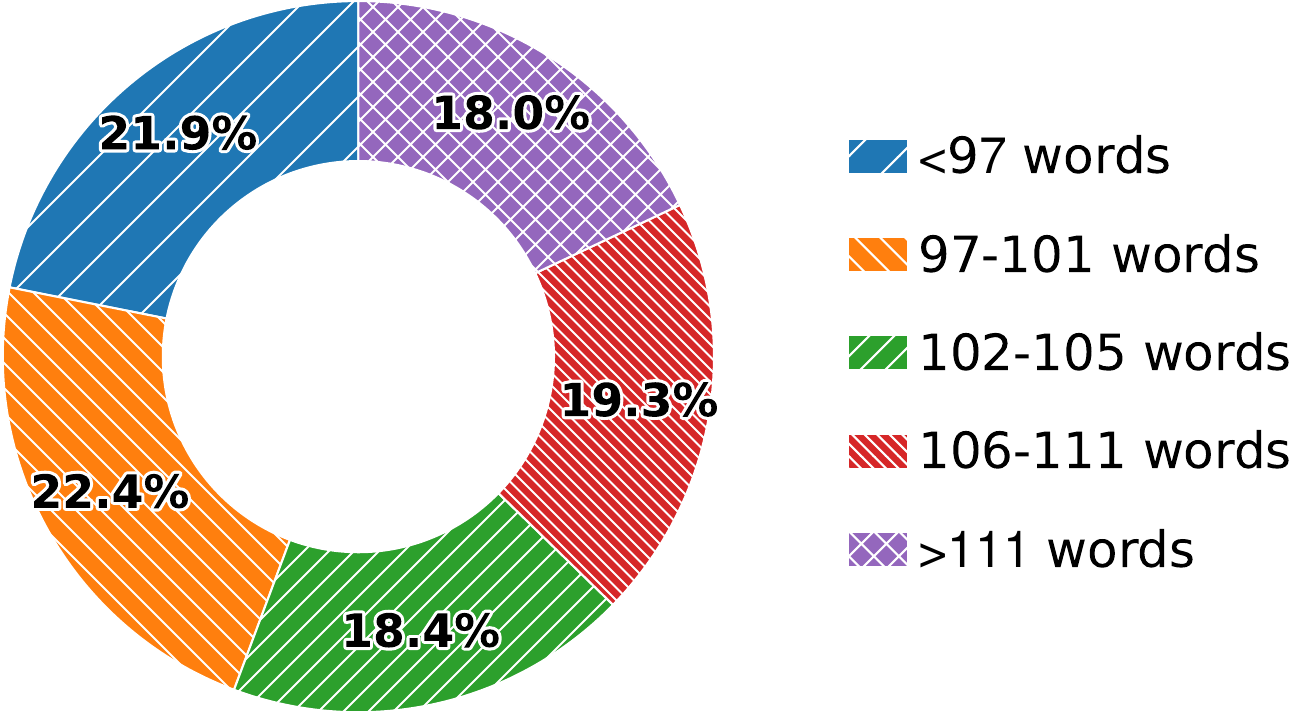}
  \end{minipage}
  \hfill 
  \begin{minipage}[b]{0.21\linewidth}
    \centering
    {\small (c) Clip Duration Donut}\par
    \vspace{0.3em}
    \includegraphics[width=\linewidth]{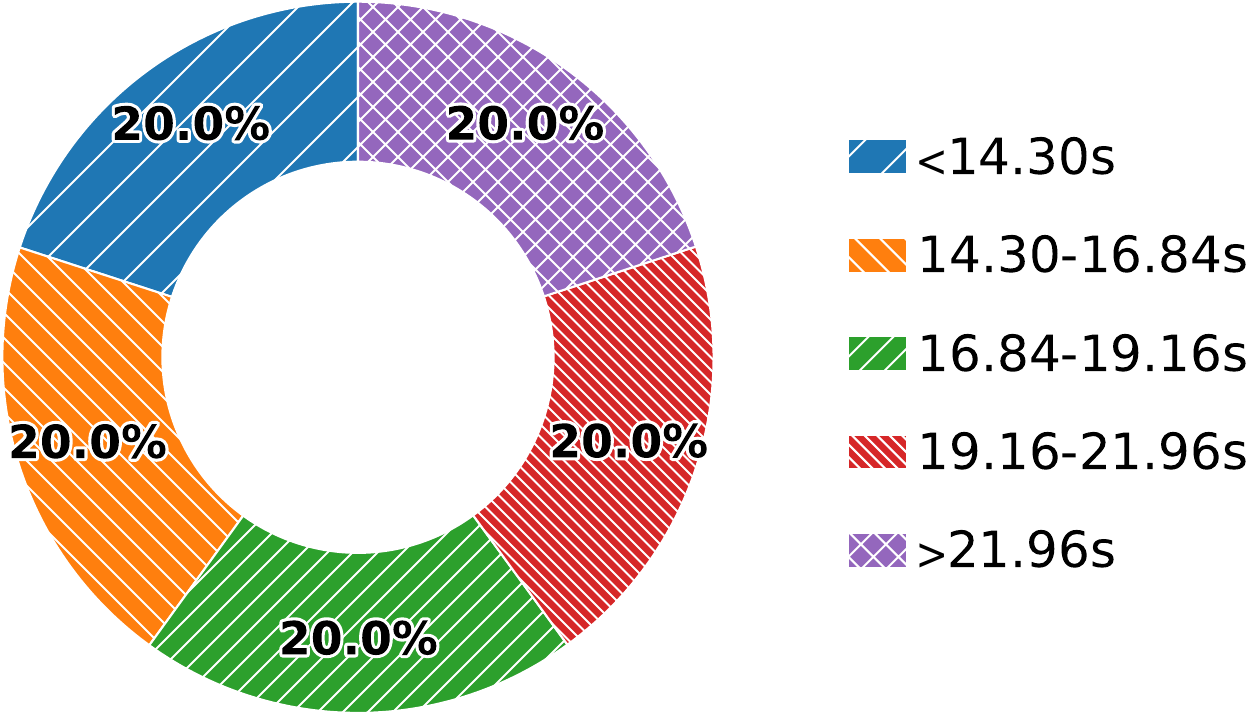}
  \end{minipage}
  \hfill 
  \begin{minipage}[b]{0.19\linewidth}
    \centering
    {\small (d) Clip Per Video Donut}\par
    \vspace{0.3em}
    \includegraphics[width=\linewidth]{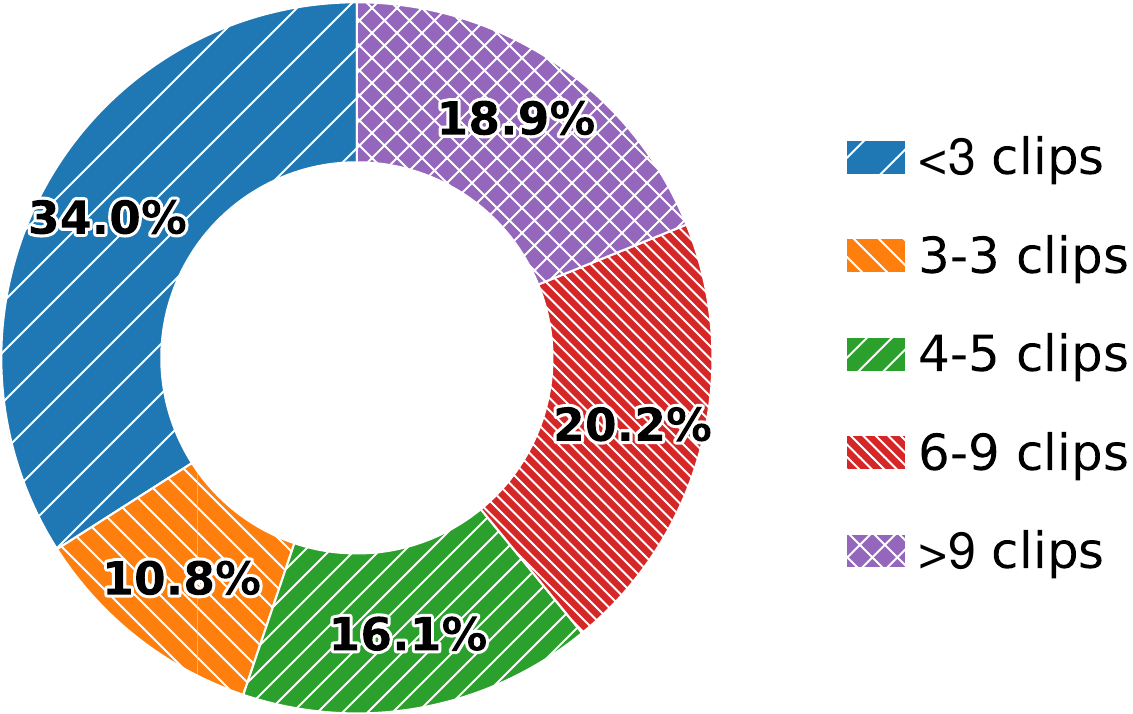}
  \end{minipage}
  
  \caption{Dataset-level statistics of the constructed benchmark. The four donut charts report the distributions of query length, reason explanation length, affective event clip duration, and the number of affective clips contained in each full video.}
  \label{fig:details}
\end{figure*}

\subsection{Data Statistics and Split}

We further analyze the statistical characteristics of the constructed dataset from several aspects, including emotion category distribution, query length, rationale length, clip duration, and the number of affective clips contained in each full video.

\textbf{(1) Emotion-Category Distribution:}
Fig.~\ref{fig:statistics} shows the distribution of emotion categories in the dataset. Overall, the dataset covers 13 emotion categories. Neutral accounts for the largest proportion, reaching 35.8\%, followed by Joy at 19.4\%. Angry and Trust also occupy relatively large proportions, accounting for 8.7\% and 8.2\%, respectively. Other affective categories, such as Sadness, Anticipation, Surprise, Fear, and Horror, appear with smaller but still non-negligible proportions. In contrast, Satisfaction, Love, Disgust, and Pride occur less frequently, reflecting the long-tailed nature of affective events in real-world videos.

\textbf{(2) Query Length Statistics:}
Fig.~\ref{fig:details} provides a detailed visualization of dataset-level attributes. Regarding query length, most queries are short and concise, which is consistent with the vague retrieval setting, where users typically describe remembered video content using incomplete or subjective expressions. Specifically, 28.7\% of queries contain fewer than 21 words, while 24.1\% fall within the range of 23--24 words. This indicates that the constructed queries are generally compact while still preserving sufficient semantic cues for retrieval.

\textbf{(3) Rationale Length:} For rationale explanations, the length distribution is relatively balanced. The proportions across different length intervals are close to one another: 21.9\% of explanations contain fewer than 97 words, 22.4\% contain 97--101 words, 18.4\% contain 102--105 words, 19.3\% contain 106--111 words, and 18.0\% contain more than 111 words. This balanced distribution suggests that the generated rationales maintain a relatively stable level of detail, providing sufficient evidence for affective understanding without being overly verbose.

\textbf{(4) Clip Duration:} The clip-duration statistics show that affective event clips are evenly distributed across five duration ranges. Each duration interval accounts for approximately 20.0\% of the data, including clips shorter than 14.30 seconds, clips between 14.30 and 16.84 seconds, clips between 16.84 and 19.16 seconds, clips between 19.16 and 21.96 seconds, and clips longer than 21.96 seconds. This balanced duration distribution helps prevent the benchmark from being biased toward either very short or very long affective events.

\textbf{(5) Number of Affective Clips per Full Video:} We also analyze the number of affective clips contained in each full video. As shown in Fig.~\ref{fig:details}, 34.0\% of full videos contain fewer than three clips, while 10.8\% contain exactly three clips. Videos containing four to five clips account for 16.1\%, and those containing six to nine clips account for 20.2\%. In addition, 18.9\% of full videos contain more than nine affective clips. These statistics indicate that the dataset includes both videos with sparse affective events and videos with dense emotional transitions, making it suitable for evaluating models under diverse temporal localization conditions.

To ensure fair evaluation and avoid data leakage, we split the dataset at the full-video level rather than at the query or clip level. In this way, affective clips and queries derived from the same full video are assigned to the same subset. Specifically, the full videos are divided into training, validation, and test sets with a ratio of \textbf{0.5, 0.25, 0.25} \cite{krishna2017dense, gao2017tall}. This split protocol ensures that models are evaluated on unseen videos, thereby providing a more reliable assessment of their generalization ability in affective video grounding and explanation generation.

\subsection{Comparison with Existing Datasets}

\begin{table*}[t]
\centering
\caption{Comparison with commonly used affective datasets. By comparing our dataset with classification-, description-, and localization-oriented benchmarks, we demonstrate that the proposed dataset is better aligned with real-world application scenarios. $I$, $A$, $V$, and $T$ denote image, audio, video, and text, respectively.}
\label{tab:dataset_comparison}
\resizebox{\textwidth}{!}{
\begin{tabular}{llcccccccc}
\toprule
Type & Dataset & Modality & \#Samples & Duration & Localization & Multi-dim.\ Explanation & Flexible Query & \#Emotions & Annotation \\
\midrule

\multirow{8}{*}{Categorical}
& RAF-DB \cite{li2017reliable}             & I       & 29,672  & \multirow{8}{*}{\centering $<1$ min} & \xmark & \xmark & \xmark & 7   & Human \\
& AffectNet \cite{mollahosseini2017affectnet}          & I       & 450,000 &                                        & \xmark & \xmark & \xmark & 8   & Human \\
& EmoDB  \cite{burkhardt2000database}             & A       & 535    &                                        & \xmark & \xmark & \xmark & 7   & Human \\
& MSP-Podcast \cite{lotfian2017building}         & A       & 73,042  &                                        & \xmark & \xmark & \xmark & 8   & Human \\
& DFEW \cite{jiang2020dfew}               & V       & 11,697  &                                        & \xmark & \xmark & \xmark & 7   & Human \\
& FERV39k \cite{wang2022ferv39k}            & V       & 38,935  &                                        & \xmark & \xmark & \xmark & 7   & Human \\
& MER2023 \cite{lian2023mer}             & A V T   & 5,030   &                                        & \xmark & \xmark & \xmark & 6   & Human \\
& MELD \cite{poria2019meld}               & A V T   & 13,708  &                                        & \xmark & \xmark & \xmark & 7   & Human \\

\midrule

\multirow{8}{*}{Descriptive}
& EmoVIT \cite{xie2024emovit}              & I       & 51,200  & \multirow{8}{*}{\centering $<1$ min} & \xmark & \xmark & \xmark & 988  & Model \\
& MERR-Coarse \cite{cheng2024emotion}         & A V T   & 28,618  &                                        & \xmark & \xmark & \xmark & 113  & Model \\
& MAFW \cite{liu2022mafw}                & A V T   & 10,045  &                                        & \xmark & \xmark & \xmark & 399  & Human \\
& OV-MERD \cite{lian2024open}            & A V T   & 332    &                                        & \xmark & \xmark & \xmark & 236  & Human-led + Model-assisted \\
& MERR-Fine \cite{cheng2024emotion}          & A V T   & 4,487   &                                        & \xmark & \xmark & \xmark & 484  & Human-led + Model-assisted \\
& MER-Caption \cite{lian2025affectgpt}         & A V T   & 115,595 &                                        & \xmark & \xmark & \xmark & 2932 & Model-led + Human-assisted \\
& MER-Caption+ \cite{lian2025affectgpt}        & A V T   & 31,327  &                                        & \xmark & \xmark & \xmark & 1972 & Model-led + Human-assisted \\
& EmoVCause \cite{huang2025emodetective}            & A V T   & 14,000  &                                        & \xmark & \cmark & \xmark & 6 or 8 & Model-led + Human-assisted \\

\midrule

\multirow{2}{*}{Localization}
& TSL300 \cite{zhang2022temporal}              & A V     & 300    & 2--19 min & \cmark & \xmark & \xmark & 2  & Human \\
& MovieGraph \cite{vicol2018moviegraphs}         & A V     & 9,311   & $<1$ min  & \cmark & \xmark & \xmark & 26 & Human \\

\midrule

\multirow{2}{*}{Grounding}
& Hawkeye \cite{zhao2024hawkeye}            & V       & 1,388   & $<3$ min & \cmark & \xmark & \xmark & 2 & Model-led + Human-assisted \\
& Omni-SILA (Explicit) \cite{luo2025omni}& V       & 77,000  & $\sim$   & \cmark & \xmark & \xmark & 3 & Model-led + Human-assisted \\

\midrule

VQAU
& VQAU-Bench          & A V T   & 48,876  & 1--30 min & \cmark & \cmark & \cmark & 13 & Model-led + Human-assisted \\

\bottomrule
\end{tabular}
}
\end{table*}

Table \ref{tab:dataset_comparison} compares VQAU-Bench with representative existing datasets across several related directions, including emotion classification, descriptive affective understanding, video localization, and affect-related video grounding. Overall, although existing datasets have advanced emotion recognition, affective description, event localization, and scene grounding, they remain insufficient for supporting the unified task studied in this paper: jointly performing affective clip localization, emotion classification, and reason analysis under vague-query guidance.

Among these datasets, categorical benchmarks are primarily designed for emotion classification on predefined clips. While they can evaluate a model’s ability to distinguish emotions, they cannot assess whether the model can discover target affective events in raw videos. Descriptive datasets further incorporate textual descriptions or explanatory annotations, thereby supporting the evaluation of fine-grained affective semantics and reasoning. However, most of them still rely on clip-level inputs, lacking temporal localization capability and flexible queries that reflect real user retrieval needs. Localization and grounding datasets have begun to address temporal localization in videos. However, the former typically do not model affective semantics or rationale explanations, while the latter, despite combining language guidance with video localization, largely rely on fixed task instructions and still fail to capture realistic scenarios in which users retrieve affective content through vague, subjective, and open-ended natural language.

\begin{figure*}[t]
  \centering
  \includegraphics[width=1.0\linewidth]{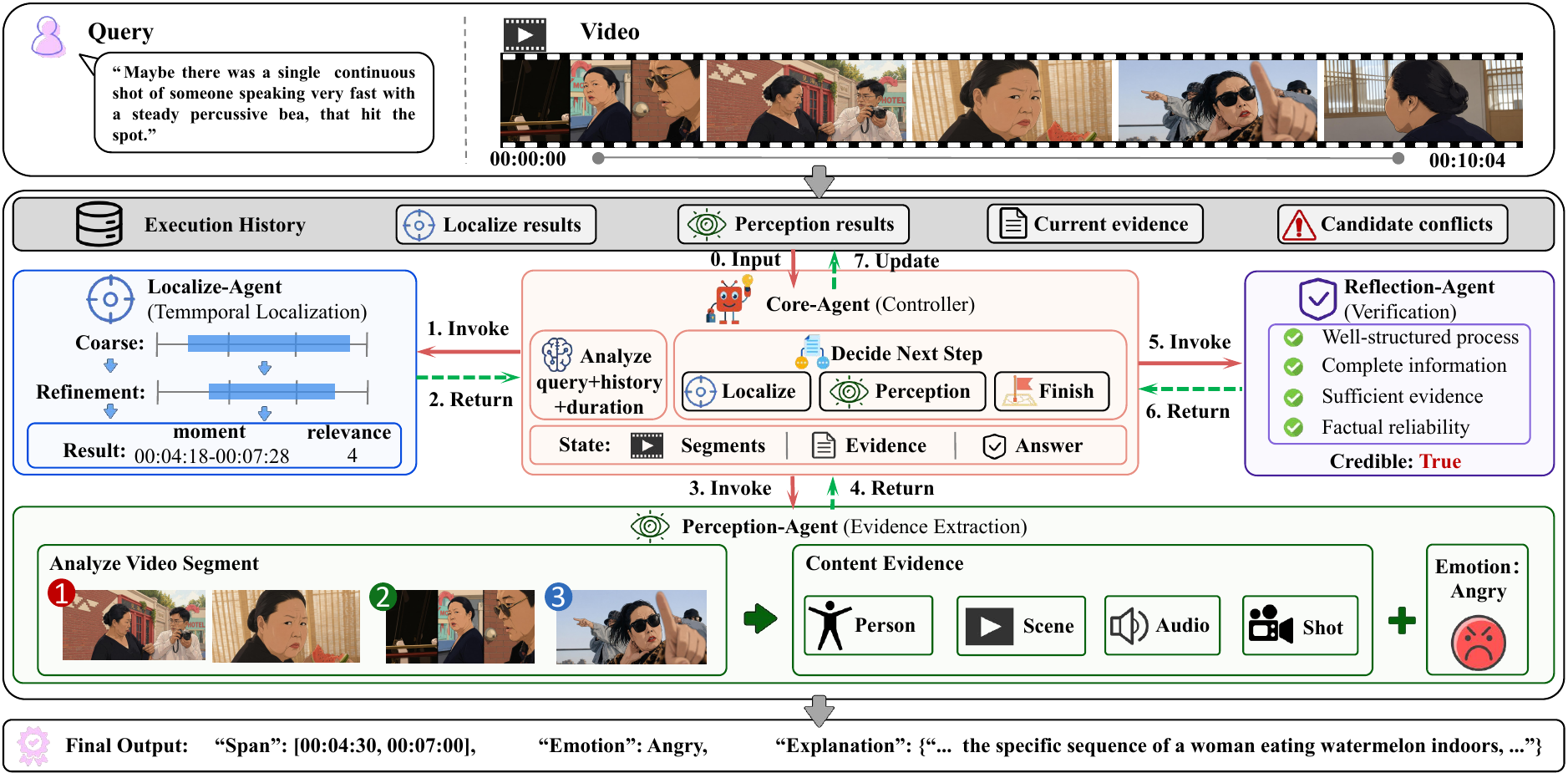}
  \caption{Overall workflow of AffectSeek framework. Given a vague query and a full video, the CoreAgent dynamically invokes the LocalizeAgent for temporal localization and the LocalizeAgent for evidence extraction. TheReflectionAgent verifies the reasoning process, information completeness, evidence sufficiency, and factual reliability before producing the final span, emotion label, and explanation.}
  \label{fig:affectagent}
\end{figure*}

In contrast, VQAU-Bench is more closely aligned with the problem studied in this paper in both task formulation and data organization. It is the only dataset in the table that simultaneously supports affective temporal localization, multi-dimensional rationale explanation, and flexible query expression within a unified benchmark. In addition, VQAU-Bench covers three modalities, namely audio, video, and text, and contains 48,876 query-clip pairs, 16,292 affective event clips, and 13 emotion categories. It therefore provides a more systematic basis for evaluating affective localization, emotion discrimination, and rationale analysis. More importantly, the rationale explanations in VQAU-Bench are not simple extensions of emotion labels. Instead, they are jointly aligned with the queries and video evidence, making the dataset better suited for evaluating model consistency and interpretability in complex affective understanding tasks.

\section{AffectSeek: A Multi-Agent Framework for Vague-Query-driven video Affective Understanding}
\label{Affect-Agent}

\subsection{Overall Framework}

Existing affective video understanding methods mainly focus on recognizing predefined emotion categories or assessing the overall affective state of a video. These methods typically assume that each input sample has an explicit emotion label and formulate the task as either classification or description generation. However, the task studied in this paper is more closely aligned with real-world application scenarios. It requires the model to understand affective queries expressed in free-form language, localize the corresponding affective events in the video, predict their emotion categories, and provide evidence-based explanations. This task involves query understanding, temporal localization, clip-level perception, affective reasoning, and evidence verification, making it difficult to address reliably through one-shot inference with a single model.

To address this challenge, we propose \textbf{AffectSeek}, a multi-agent framework for affective video understanding guided by ambiguous descriptions. The detailed workflow is illustrated in Fig. \ref{fig:affectagent}. Inspired by cognitive psychology, which decomposes human cognition into processes such as perception, attention, reasoning, language, and decision-making \cite{lake2017building}, AffectSeek breaks this complex task into functional subprocesses that are collaboratively handled by Localize-Agent, Perception-Agent, Reflection-Agent, and Core-Agent. Through inter-agent collaboration, the framework performs evidence-based reasoning in a stepwise manner. Specifically, \textbf{Localize-Agent} invokes the \textit{localization tool} and \textit{refinement tool} to identify query-relevant candidate clips and refine their temporal boundaries. \textbf{Perception-Agent} invokes the \textit{verification tool} and \textit{summary tool} to analyze the candidate clips, predict emotion categories, and extract supporting evidence. \textbf{Reflection-Agent} verifies whether the candidate clips, emotion judgments, and explanations are sufficiently supported by evidence. \textbf{Core-Agent} coordinates the overall reasoning process. During inference, Core-Agent dynamically invokes the appropriate agents according to the current evidence state and generates the final answer only after sufficient evidence has been verified. The detailed prompts are provided in the Appendix B.

\subsection{Functional Agent Decomposition}
\subsubsection{\textbf{Core-Agent}}
Core-Agent serves as the central orchestrator of the Agent framework, coordinating agent invocation and guiding the overall reasoning process. Specifically, it dynamically makes decisions based on the current evidence state, including whether to further invoke Localize-Agent to expand or refine candidate clips, whether to invoke Perception-Agent to obtain fine-grained affective evidence, whether the available evidence is sufficient to support the emotion judgment, and whether Reflection-Agent should be introduced to verify the consistency and reliability of the current results.
Once the termination condition is satisfied, Core-Agent stops invoking additional agents. It then aggregates the collected evidence, including span evidence, emotion evidence, and key-frame evidence, and generates the final answer accordingly.

\subsubsection{\textbf{Localize-Agent}}
The Localize-Agent is responsible for identifying query-relevant temporal clips from the full video $V$ according to the input query $q$. It is equipped with three tools: \underline{\textit{localize tool, refinement tool, and finish}}. Its workflow follows a coarse-to-fine two-stage temporal localization process.

\textbf{First}, the localize tool performs coarse-grained localization over the full video $V$ based on the query $q$, retrieving a set of potentially relevant candidate clips:
\begin{equation}
\mathcal{C}=\text{LocalizeTool}(V,q)={c_i}_{i=1}^{N},
\end{equation}
where $\mathcal{C}$ denotes the candidate clip set obtained from coarse localization, and $c_i$ denotes the $i$-th candidate clip.

\textbf{Second}, the refinement tool further refines the coarse localization results $\mathcal{C}$ through fine-grained retrieval and boundary adjustment, producing a set of more precise temporal clips:
\begin{equation}
\mathcal{S}=\text{RefinementTool}(V,q,\mathcal{C})={s_i}_{i=1}^{K},
\end{equation}
where $s_i=(t_i^s,t_i^e)$ denotes the $i$-th refined clip, and $t_i^s$ and $t_i^e$ denote its start and end times, respectively. After obtaining the refined temporal clips, the finish tool terminates the localization process and returns $\mathcal{S}$ to Core-Agent for subsequent perception and verification.

During execution, Localize-Agent selects tools according to the current localization state. If no candidate clip has been obtained, it invokes the localize tool for global coarse localization. If candidate clips are available but their temporal boundaries remain imprecise, it invokes the refinement tool for fine-grained localization. Once reliable refined localization results $\mathcal{S}$ are obtained, Localize-Agent invokes finish to terminate the process and returns $\mathcal{S}$ to Core-Agent as input for the subsequent perception and verification stages.

\subsubsection{\textbf{Perception-Agent}}
This agent performs clip-level affective perception on the candidate intervals returned by Localize-Agent. \textbf{First}, given a set of candidate clips $\mathcal{S}={s_i}_{i=1}^{K}$, it verifies their relevance to the query $q$ using the \underline{\textit{verification tool}}. Unlike temporal localization, this stage focuses on evidence-level validation rather than candidate retrieval. For each candidate clip $s_i$, Perception-Agent obtains a matching score and the corresponding perceptual evidence:
\begin{equation}
(\alpha_i, v_i, u_i, r_i)=\text{VerificationTool}(V,q,s_i),
\end{equation}
where $\alpha_i$ denotes the query-clip matching score, $v_i$ denotes the matched visual evidence, $u_i$ denotes the uncertainty, and $r_i$ denotes the selection rationale.

\textbf{Based on the matching scores}, Perception-Agent retains candidates with sufficient relevance and merges temporally adjacent or overlapping intervals:
\begin{equation}\mathcal{S}^{*}=\text{Merge}\left({s_i \mid \alpha_i \geq \tau}\right),\end{equation}
where $\tau$ denotes the relevance threshold. This operation filters out weakly related clips while preserving continuous affective evidence across neighboring intervals.

\textbf{After the final interval $\mathcal{S}^{}$ is determined}, Perception-Agent performs clip-level summarization and emotion recognition using the \underline{\textit{summary tool}}:
\begin{equation}
(e,z)=\text{SummaryTool}(V,q,\mathcal{S}^{}),
\end{equation}
where $e$ denotes the predicted emotion category and $z$ denotes the clip summary together with the supporting affective evidence. The output of Perception-Agent includes the verified interval, the clip-level summary, the predicted emotion category, and the corresponding evidence, which are then passed to Reflection-Agent for reliability verification.

\subsubsection{\textbf{Reflection-Agent}}

Unlike Localize-Agent and Perception-Agent, Reflection-Agent does not perform temporal localization, content perception, or answer generation. Instead, it reviews the complete reasoning history to determine whether the final answer is sufficiently supported by the collected evidence.

Given the historical operation record $H$, the query $q$, and the candidate answer $a$ generated by Core-Agent, Reflection-Agent evaluates whether the agent invocation is appropriate, whether key information is complete, whether the intermediate results are consistent with the final answer, whether the answer is sufficiently supported by evidence, and whether it contains fabrication or over-inference. This process is formulated as:

\begin{equation}(c,\gamma)=\text{ReflectionAgent}(H,q,a),\end{equation}

where $c\in{\text{true},\text{false}}$ indicates whether the final answer is credible, and $\gamma$ denotes the explanation returned when the answer is judged not credible.

Specifically, Reflection-Agent checks whether Core-Agent appropriately invokes Localize-Agent and Perception-Agent, whether the history contains sufficient temporal clips, emotion evidence, and key visual cues, whether the reasoning process is consistent, and whether the final answer strictly follows the historical evidence. Reflection-Agent judges the answer as credible only when the process is reasonable, the information is complete, the reasoning is consistent, and the evidence is sufficient. Otherwise, it returns the corresponding reason for the credibility failure, thereby reducing unsupported conclusions and hallucinated explanations.

\section{Evaluation Benchmark}

\subsubsection{\textbf{Localization and Classification Accuracy}}

To evaluate affective video understanding, we assess temporal localization, emotion classification, and their joint accuracy.

For temporal localization, we adopt standard metrics widely used in video grounding, including temporal Intersection over Union (tIoU), mean tIoU, and Recall@1 at different tIoU thresholds. Given a ground-truth temporal clip $g_i = (t_i^s, t_i^e)$ and a predicted clip $\hat{g}_i = (\hat{t}_i^s, \hat{t}_i^e)$ \cite{krishna2017dense, gao2017tall}, tIoU is defined as:
\begin{equation}
\text{tIoU}_i = \frac{\left| g_i \cap \hat{g}_i \right|}{\left| g_i \cup \hat{g}_i \right|}.
\end{equation}

Mean tIoU is computed by averaging the tIoU scores over all valid matched samples. We also report Recall@1 at a threshold $\tau$, which measures whether the top-1 predicted clip satisfies $\text{tIoU}_i \geq \tau$.

To jointly evaluate temporal localization accuracy (tIoU) and emotion classification accuracy (Acc), we introduce a stricter metric, Joint@$\tau$. A prediction is considered correct only when the predicted emotion label matches the ground truth and the predicted clip satisfies the specified tIoU threshold:
\begin{equation}
\text{Joint@}\tau = \frac{1}{N} \sum_{i=1}^{N} \mathbf{1}\left(\text{tIoU}_i \geq \tau \ \wedge\ e_i = \hat{e}_i \right),
\end{equation}
where $e_i$ and $\hat{e}_i$ denote the ground-truth and predicted emotion labels, respectively.
We report results at $\tau = 0.3, 0.5, 0.7$. Compared with localization or classification metrics alone, Joint@$\tau$ provides a more rigorous evaluation, as it requires both accurate temporal grounding and correct emotion recognition.

\begin{table*}[t]
    \centering
    \caption{Performance comparison with commonly used LLMs on the VQAU task. The results show that a single end-to-end model struggles to solve this task effectively, highlighting the effectiveness of the proposed AffectSeek framework. Here, IoU measures localization accuracy, while Joint evaluates performance when both temporal localization and emotion classification are correct.}
    \label{tab:sota}
    \small
    \setlength{\tabcolsep}{3.2pt}
    \renewcommand{\arraystretch}{1.08}
    \begin{tabular}{lcccccccc}
    \toprule
        \multicolumn{1}{c}{\multirow[c]{2}{*}{Model}}
        & \multirow[c]{2}{*}{mIoU}
        & \multicolumn{3}{c}{Localization R@1}
        & \multicolumn{3}{c}{Joint R@1}
        & \multirow[c]{2}{*}{Score} \\
        \cmidrule(lr){3-5} \cmidrule(lr){6-8}
        &
        & tIoU$\geq$0.3
        & tIoU$\geq$0.5
        & tIoU$\geq$0.7
        & tIoU$\geq$0.3
        & tIoU$\geq$0.5
        & tIoU$\geq$0.7
        & \\
    \midrule
        \rowcolor{gray!20} \emph{Fine-tuned Models} & ~ & ~ & ~ & ~ & ~ & ~ & ~ & ~ \\ 
        AffectGPT \cite{lian2025affectgpt} 
        & 5.22 & 7.85 & 3.59 & 1.06 & 3.96 & 1.89 & 0.67 & 3.50 \\ 
        
        Emotion-LLaMa \cite{cheng2024emotion} 
        & 12.95 & 8.57 & 3.29 & 1.99 & 3.79 & 1.55 & 0.11 & 4.52 \\ 
        
        Emotion-LLaMa-v2 \cite{peng2026emotion} 
        & 8.91 & 13.79 & 5.62 & 2.40 & 7.10 & 2.76 & 1.28 & 3.25 \\ 
    \midrule
        \rowcolor{gray!20} \emph{Commercial Models} & ~ & ~ & ~ & ~ & ~ & ~ & ~ & ~\\ 
        Qwen3.5 \cite{qwen35blog}  
        & 9.02 & 12.90 & 5.97 & 2.16 & 9.60 & 4.39 & 1.52 & 4.90 \\ 
        
        Gemini3.1-Pro \cite{team2023gemini} 
        & 10.61 & 14.88 & 6.81 & 2.63 & 8.14 & 3.72 & \textbf{1.48} & 5.12 \\ 
        
        GPT-5.4 \cite{openai2026gpt54}  
        & 6.26 & 8.22 & 3.88 & 1.54 & 4.62 & 2.14 & 0.86 & 5.15 \\ 
        
        Kimi-2.5 \cite{team2026kimi}  
        & 6.68 & 7.84 & 3.94 & 1.62 & 3.93 & 1.87 & 0.77 & 3.02 \\ 
    \midrule
        \rowcolor{gray!20} \emph{Multi-Agents} & ~ & ~ & ~ & ~ & ~ & ~ & ~ & ~\\ 
        AffectSeek (Ours)
        & \textbf{19.55} & \textbf{33.33} & \textbf{14.44} & \textbf{3.33} 
        & \textbf{15.91} & \textbf{4.55} & 1.14 & \textbf{7.09} \\ 
    \bottomrule
    \end{tabular}
\end{table*}

\subsubsection{\textbf{Explanation Faithfulness Evaluation}}

In addition to prediction accuracy, we evaluate the correctness and faithfulness of the generated explanations. Since explanation quality involves multimodal reasoning and cannot be reliably captured by simple matching metrics, we adopt an LLM-as-a-Judge paradigm.

Specifically, we use a strong large language model (GPT-5.4-Pro) as the evaluator to assess whether each generated explanation is grounded in the reference evidence. The evaluation focuses on correctness rather than fluency, style, or verbosity. Each explanation is scored along six dimensions.

\begin{tcolorbox}[
    colback=gray!10,
    colframe=gray!50,
    boxrule=0.5pt,
    arc=2pt,
    before upper={\setlength{\parindent}{0pt}}
]
\noindent\textbf{Scoring dimensions:}
\begin{enumerate}[leftmargin=1.5em, itemsep=0.2em, topsep=0.3em]
    \item Sentence-level semantic consistency
    \item Visual expressive evidence
    \item Vocal prosody evidence
    \item Event or story content
    \item Cinematographic evidence
    \item Hallucination
\end{enumerate}
\end{tcolorbox}

Each dimension is scored on a scale of ${0,1,2}$, where 2 denotes a correct and well-supported assessment, 1 denotes a partially correct or weakly supported assessment, and 0 denotes an incorrect or unsupported assessment.

The total score is computed as:
\begin{equation}
\text{Score} = \sum_{k=1}^{6} s_k,
\end{equation}
where $s_k$ denotes the score of the $k$-th dimension. Based on the total score, each explanation is assigned to one of three quality levels: \textbf{Correct} for scores in $[10-12]$, \textbf{Partially Correct} for scores in $[6-10)$, and \textbf{Incorrect} for scores in $[0-6)$.

To ensure strict evaluation, we further introduce hard constraints. If an explanation contains a major hallucination, such as fabricated evidence or incorrect causal attribution, it is directly classified as incorrect regardless of its total score. This protocol emphasizes evidence grounding and penalizes unsupported or speculative reasoning, thereby enabling a more reliable assessment of explanation quality in affective video understanding. The detailed prompts are provided in the Appendix C.

\begin{figure*}[t]
  \centering
  \includegraphics[width=1.0\linewidth]{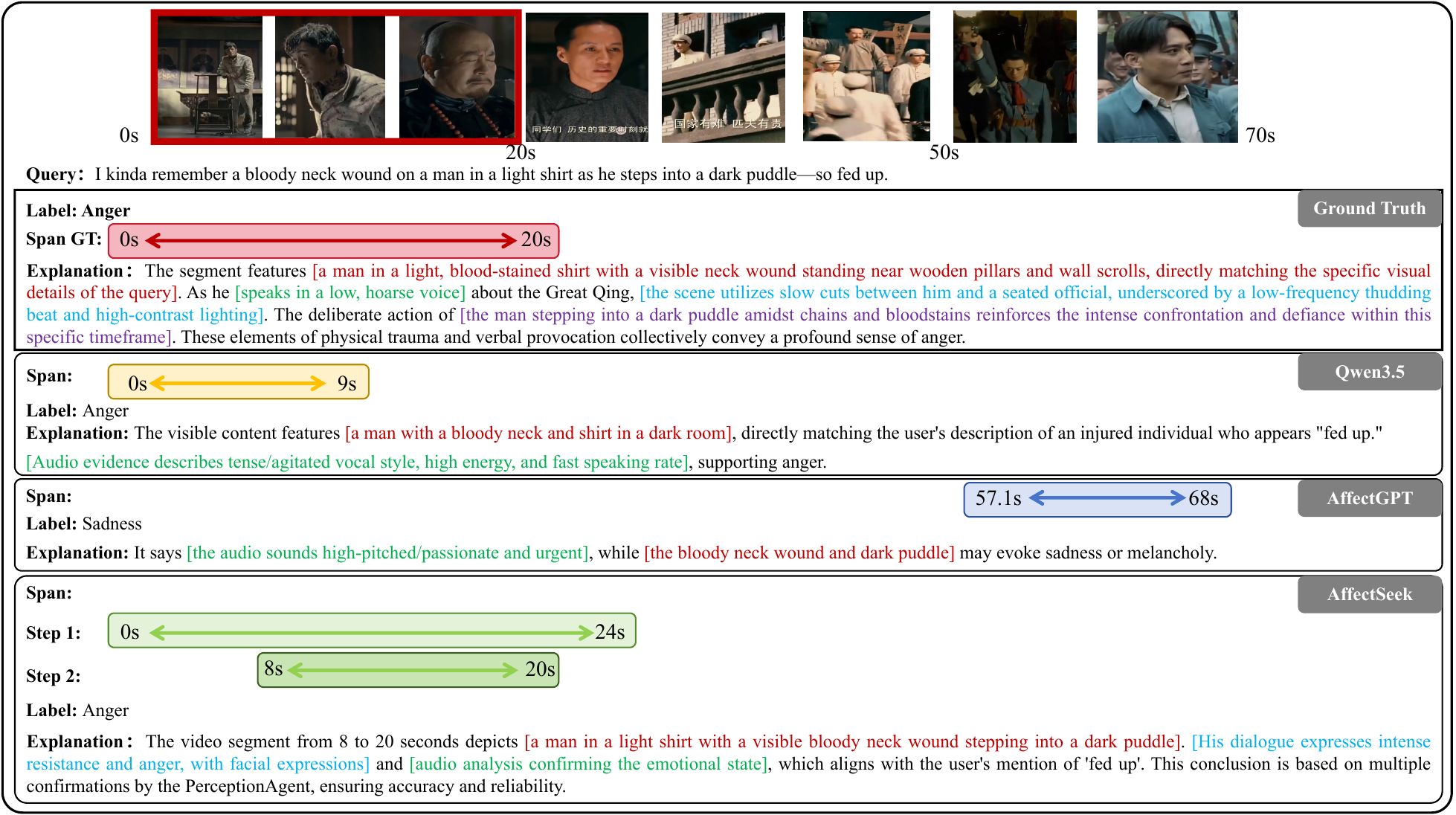}
  \caption{Visualization of representative prediction results. The comparison shows that AffectSeek effectively localizes affective event clips, identifies the emotions they contain, and generates more complete rationale explanations than competing methods.}
  \label{fig:results}
\end{figure*}

\section{Results and Discussion}\label{results}

\subsection{Comparison with commonly emotion LLMs}
Table~\ref{tab:sota} compares AffectSeek with fine-tuned affective models and commercial large multimodal models on the VQAU task. We evaluate temporal localization using mIoU and Localization R@1, joint localization-classification performance using Joint R@1, and the truthfulness and validity of generated rationales using an LLM-as-a-Judge protocol.

The results show that single end-to-end models achieve limited performance on VQAU. Fine-tuned affective models, including AffectGPT, Emotion-LLaMA, and Emotion-LLaMA-v2, obtain relatively low localization and joint scores, indicating that emotion recognition on given inputs is insufficient for vague-query-based full-video understanding. Commercial models, such as Gemini3.1-Pro, GPT-5.4, and Kimi-2.5, outperform most fine-tuned models, but they still struggle to simultaneously localize affective clips and predict their emotion categories.

In contrast, AffectSeek achieves the best performance on most metrics. It obtains an mIoU of 19.55, clearly outperforming the strongest fine-tuned baseline, Emotion-LLaMA, as well as the strongest commercial baseline, Gemini3.1-Pro. It also achieves the highest Localization R@1 scores across all tIoU thresholds, demonstrating a stronger ability to identify affective clips from full videos based on vague queries. In the joint evaluation, AffectSeek achieves the best Joint R@1 scores at tIoU$\geq$0.3 and tIoU$\geq$0.5, and obtains the highest rationale explanation score of 7.09. These results demonstrate the effectiveness of decomposing VQAU into multiple collaborative stages, including query understanding, candidate localization, clip verification, emotion reasoning, and rationale generation.

AffectSeek performs slightly worse than Gemini3.1-Pro on Joint R@1 at tIoU$\geq$0.7. A possible reason is that the multi-agent process tends to preserve the complete affective event span in order to retain sufficient contextual evidence for emotion reasoning and explanation. While this strategy improves performance under moderate thresholds, it can be penalized by the strictest boundary-sensitive metric when the predicted segment is slightly wider than the ground-truth annotation. Overall, AffectSeek shows clear advantages in localization, joint evaluation, and overall performance, confirming the effectiveness of the proposed multi-agent framework for the VQAU task.

\begin{figure}[t]
  \centering
  \includegraphics[width=1.0\linewidth]{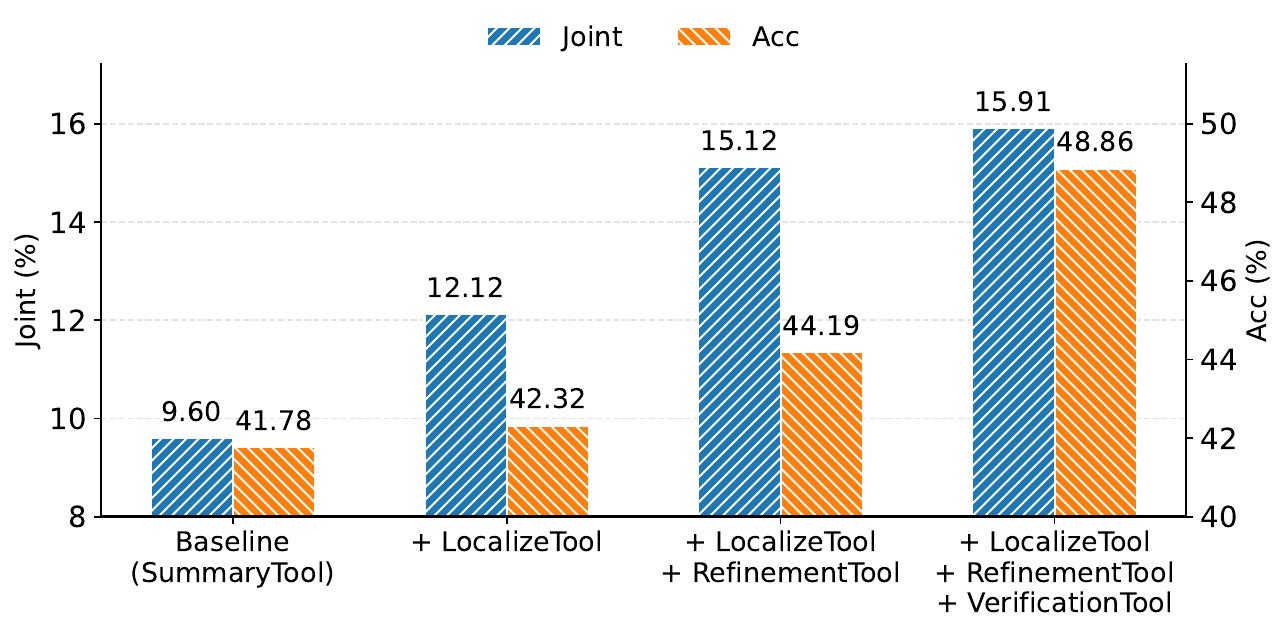}
  \caption{Ablation Study of Individual Tools in AffectSeek. The comparison shows that fine-grained localization and verification tools effectively improve the agent’s ability to perform temporal localization and emotion classification.}
  \label{fig:ablation_tool}
\end{figure}

Meanwhile, we further analyze the outputs of different methods. As shown in Fig. \ref{results}, Qwen3.5 predicts the correct emotion category, but its localized clip only covers 0--9 s. This prediction mainly captures the local visual cue of the character’s injury, while failing to include the subsequent key affective event, namely the character stepping into a dark puddle. In contrast, AffectGPT localizes the clip to 57.1--68 s, which clearly deviates from the annotated segment, and incorrectly predicts the emotion as Sadness. These results suggest that single models may suffer from semantic matching errors and affective interpretation biases when handling vague affective queries.
By comparison, AffectSeek first identifies a coarse temporal range of 0--24 s and then refines it to 8--20 s, enabling more accurate coverage of the main affective event and correctly predicting the emotion category as Anger. More importantly, the rationale generated by AffectSeek integrates visual evidence, including bloodstains, the neck wound, and the action of stepping into a dark puddle, together with multimodal cues such as the character’s dialogue, facial expressions, and audio-based emotional state. Therefore, AffectSeek provides a more comprehensive explanation for both the localization result and the emotion judgment. This further demonstrates its effectiveness in vague query understanding, affective clip localization, and multimodal rationale generation.


\subsection{Tool-Level Ablation Study}
To verify the effectiveness of each tool in AffectSeek, we conduct a tool-level ablation study, as shown in Fig.~\ref{fig:ablation_tool}. Since this study aims to assess the contribution of each tool from two perspectives, namely the overall ability to solve the task and the accuracy of affective understanding, we use Joint and Acc as the main evaluation metrics. The results show that model performance improves steadily as LocalizeTool, RefinementTool, and VerificationTool are progressively introduced. After adding LocalizeTool, Joint increases to 12.12, indicating that coarse localization helps identify relevant affective regions. With the further addition of RefinementTool, Joint rises to 15.12, suggesting that fine-grained retrieval improves target segment localization. Finally, after incorporating VerificationTool, Joint and Acc reach 15.91 and 48.86, respectively. These results demonstrate that candidate segment verification further enhances the reliability of both localization and emotion prediction.


\subsection{Agent-Level Ablation Study}
\begin{figure}[t]
  \centering
  \includegraphics[width=1.0\linewidth]{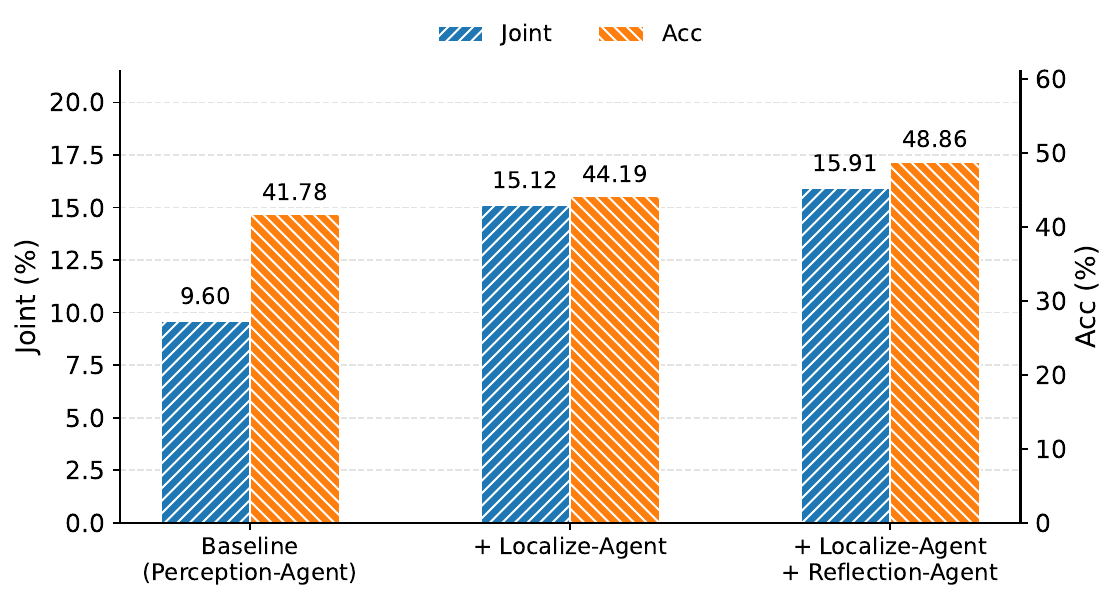}
  \caption{Ablation study on Each Agent in AffectSeek. The experimental results demonstrate the effectiveness of the fine-grained localization performed by Localize-Agent and the reflection mechanism introduced by Reflection-Agent.}
  \label{fig:ablation_agent}
\end{figure}

Following the same evaluation protocol, we further conduct an agent-level ablation study, as shown in Fig.~\ref{fig:ablation_agent}. When only Perception-Agent is used, the model achieves relatively low performance, indicating that a single perception module is insufficient for vague affective localization in full videos. After introducing Localize-Agent, both localization and emotion recognition improve, demonstrating its effectiveness in narrowing the search space and providing reliable candidate segments. With the further addition of Reflection-Agent, performance continues to improve, suggesting that consistency checking and evidence verification help reduce incorrect predictions. Overall, the tool-level and agent-level ablation studies demonstrate the effectiveness of the multi-tool collaboration and multi-agent cooperation design in AffectSeek.

\section{Conclusion}
\label{conclusion}

This paper studies affective understanding beyond traditional clip-level emotion classification. Rather than assuming that target affective clips are pre-specified, we focus on discovering, recognizing, and explaining affective moments in long videos guided by vague user queries. This setting better reflects practical multimedia interaction scenarios, where users often express their needs through subjective and imprecise natural language instead of precise timestamps or standardized emotion labels. To support this setting, we introduce VQAU, a task that jointly models affective moment localization, emotion classification, and evidence-grounded rationale generation. In this way, affective understanding is extended from passive recognition of given content to user-centered, query-driven, and interpretable affective reasoning.

This work also represents an initial step toward agentic affective understanding. Although VQAU-Bench provides a complete query-clip-emotion-rationale annotation chain, it is still limited in scenario coverage, query diversity, and the representation of complex emotional expressions. In open-world environments, user descriptions can be fragmented and diverse, while emotional interpretation is inherently subjective and context-dependent. These factors create a semantic and interaction gap between the current benchmark and real-world deployment. Future work will expand VQAU-Bench, develop more robust agentic reasoning frameworks, and explore interactive mechanisms to better align vague user intent with multimodal affective evidence.

%

\bibliographystyle{IEEEtran}
\bibliography{ref}


 




\vfill

\end{document}